\documentclass[10pt,twocolumn,letterpaper]{article}

\usepackage{cvpr}              

\usepackage{multirow}
\usepackage[table,xcdraw]{xcolor}
\usepackage{xcolor}
\usepackage{graphicx}
\usepackage{amsmath}
\usepackage{amssymb}
\usepackage{enumitem}
\usepackage{booktabs}
\usepackage{times}
\usepackage{epsfig}
\usepackage{graphicx}
\usepackage{makecell}
\usepackage{amsmath}
\usepackage{amssymb}
\usepackage[font={small}]{caption}
\usepackage{diagbox}
\usepackage{ifthen}
\usepackage{bm}
\usepackage{balance}
\usepackage{lipsum}
\usepackage{booktabs}
\usepackage{verbatim}
\usepackage{graphicx}
\usepackage{color}
\usepackage{bbding}
\usepackage{pifont}
\usepackage{wasysym}
\usepackage{amssymb}
\usepackage{diagbox}
\usepackage{ifthen}
\usepackage{bbding} 
\usepackage{bm}
\usepackage{threeparttable}
\usepackage{lipsum}
\usepackage{animate}
\usepackage{rotating} 
\usepackage{color}
\usepackage{makecell}
\usepackage{amsmath}
\usepackage{bbding}
\usepackage{pifont}
\usepackage{lipsum}
\usepackage{algorithmicx}
\usepackage{algorithm}
\usepackage{algpseudocode}
\usepackage{amsmath}  
\usepackage[misc]{ifsym}
\usepackage{url}

\usepackage[utf8]{inputenc} 

\usepackage{fontawesome}
\usepackage[accsupp]{axessibility}

\definecolor{cvprblue}{rgb}{0.21,0.49,0.74}
\usepackage[pagebackref,breaklinks,colorlinks,allcolors=cvprblue]{hyperref}

\usepackage[capitalize]{cleveref}
\crefname{section}{Sec.}{Secs.}
\Crefname{section}{Section}{Sections}
\Crefname{table}{Table}{Tables}
\crefname{table}{Tab.}{Tabs.}

\def\cvprPaperID{1838} 
\def\confName{CVPR}
\def\confYear{2025}

\begin{document}

\title{PolarFree: Polarization-based Reflection-Free Imaging}

\author{Mingde Yao$^1$,~~~Menglu Wang$^3$,~~~King-Man Tam$^{1,4}$,~~~Lingen Li$^1$,~~~Tianfan Xue\textsuperscript{\faEnvelopeO}$^{1,2}$,~~~Jinwei Gu$^1$\\
$^1$The Chinese University of Hong Kong,~~ $^2$Shanghai AI Laboratory\\$^3$University of Science and Technology of China,~~$^4$Institute of Science Tokyo\\
{\tt\small mingdeyao@foxmail.com,~~tfxue@ie.cuhk.edu.hk}
}
\maketitle
\newcommand\blfootnote[1]{%
\begingroup
\renewcommand\thefootnote{}\footnote{#1}
\addtocounter{footnote}{-1}%
\endgroup
}

\blfootnote{\textsuperscript{\faEnvelopeO}Corresponding author.}	

\vspace{-0.6cm}
\begin{abstract}
\vspace{-0.4cm}
Reflection removal is challenging due to complex light interactions, where reflections obscure important details and hinder scene understanding. Polarization naturally provides a powerful cue to distinguish between reflected and transmitted light, enabling more accurate reflection removal. However, existing methods often rely on small-scale or synthetic datasets, which fail to capture the diversity and complexity of real-world scenarios.
To this end, we construct a large-scale dataset, \textit{\textbf{PolaRGB}}, for Polarization-based reflection removal of RGB images, which enables us to train models that generalize effectively across a wide range of real-world scenarios. The PolaRGB dataset contains 6,500 well-aligned mixed-transmission image pairs, 8$\times$ larger than existing polarization datasets, and is the first to include both RGB and polarization images captured across diverse indoor and outdoor environments with varying lighting conditions. Besides, to fully exploit the potential of polarization cues for reflection removal, we introduce \textit{\textbf{PolarFree}}, which leverages diffusion process to generate reflection-free cues for accurate reflection removal.
Extensive experiments show that PolarFree significantly enhances image clarity in challenging reflective scenarios, setting a new benchmark for polarized imaging and reflection removal. Code and dataset are available at \url{https://github.com/mdyao/PolarFree}.
\end{abstract}

\vspace{-0.6cm}
\label{sct:Introduction}
\section{Introduction}
\vspace{-0.1cm}

Reflection removal algorithms~\cite{tan2005separating,schechner2000polarization,agrawal2005removing,wei2019single,lei2023robust,xue2015computational} remove unwanted reflections in captured images, playing a critical role in applications such as autonomous driving~\cite{huang2016dust} and photography~\cite{yun2018reflection,hu2023single}.
This problem commonly arises when imaging through semi-reflectors, like windows or glass, and overlapping reflections may obscure important details of scenes we want to capture. This problem is often formulated~\cite{hu2021trash,hu2023single} as a linear combination of   the transmission layer $T$ and the reflection layer $R$: 
\begin{equation}~\label{eq:eq1}
\abovedisplayskip=1pt  
\abovedisplayshortskip=3pt
    M=\alpha_t  T+\alpha_r  R,
\belowdisplayskip=3pt  
\belowdisplayshortskip=1pt
\end{equation}
where $M$ is the mixed captured image, and $\alpha_t, \alpha_r$ are blending coefficients resulting from light attenuation. 

\begin{figure}[!t]
    \centering
    \includegraphics[width=1\linewidth]{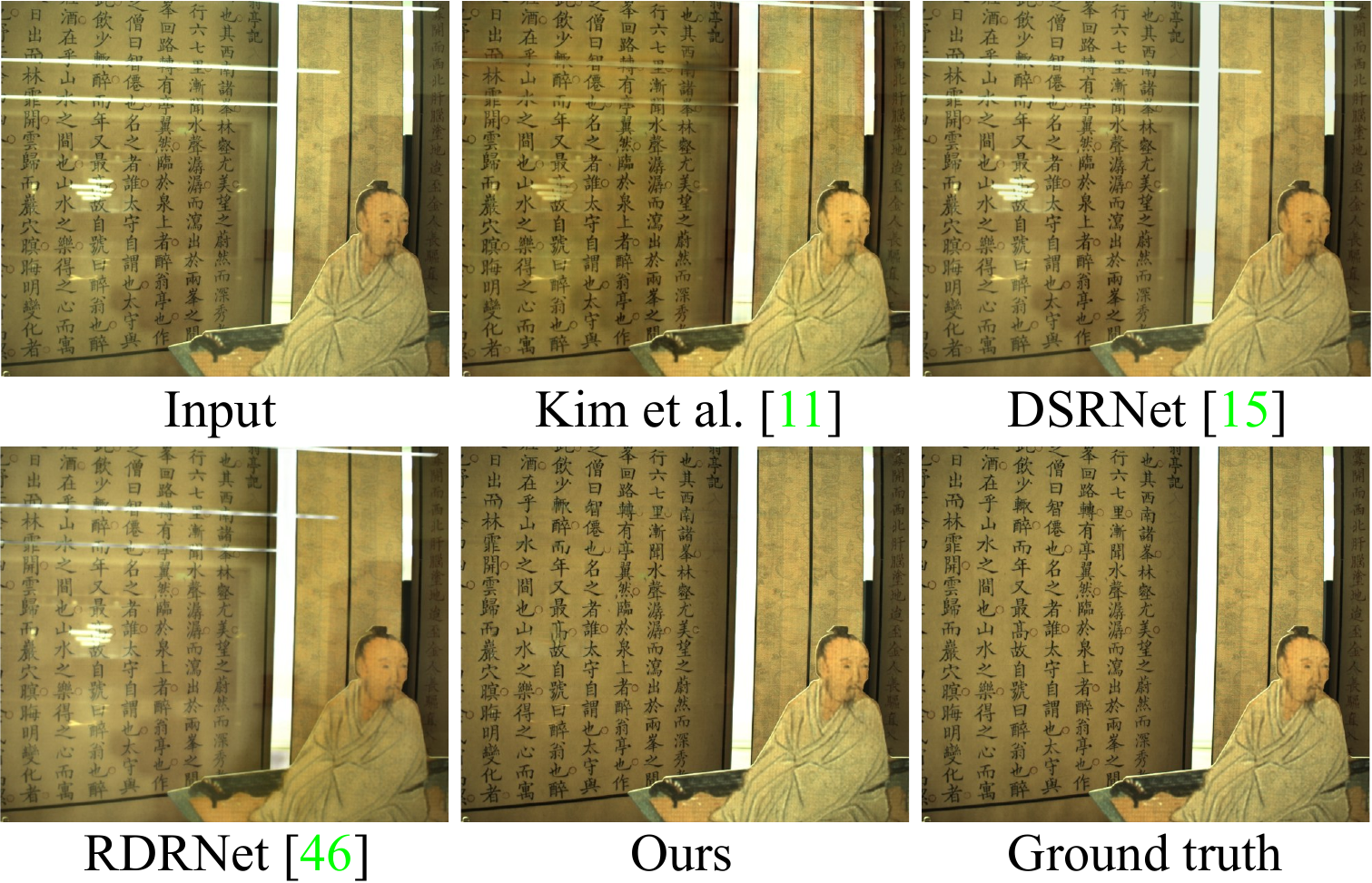}
    	\vspace{-0.7cm}
	\caption{Our PolarFree effectively leverages polarization information to remove reflections, achieving superior performance in challenging scenes with complex backgrounds and highlights where previous methods~\cite{kim2020single,hu2023single,zhu2024revisiting} often fail.}
	\label{fig:fig1}
	\vspace{-0.65cm}
\end{figure}

\begin{figure}[!t]
	\centering
	\includegraphics[width=1\linewidth]{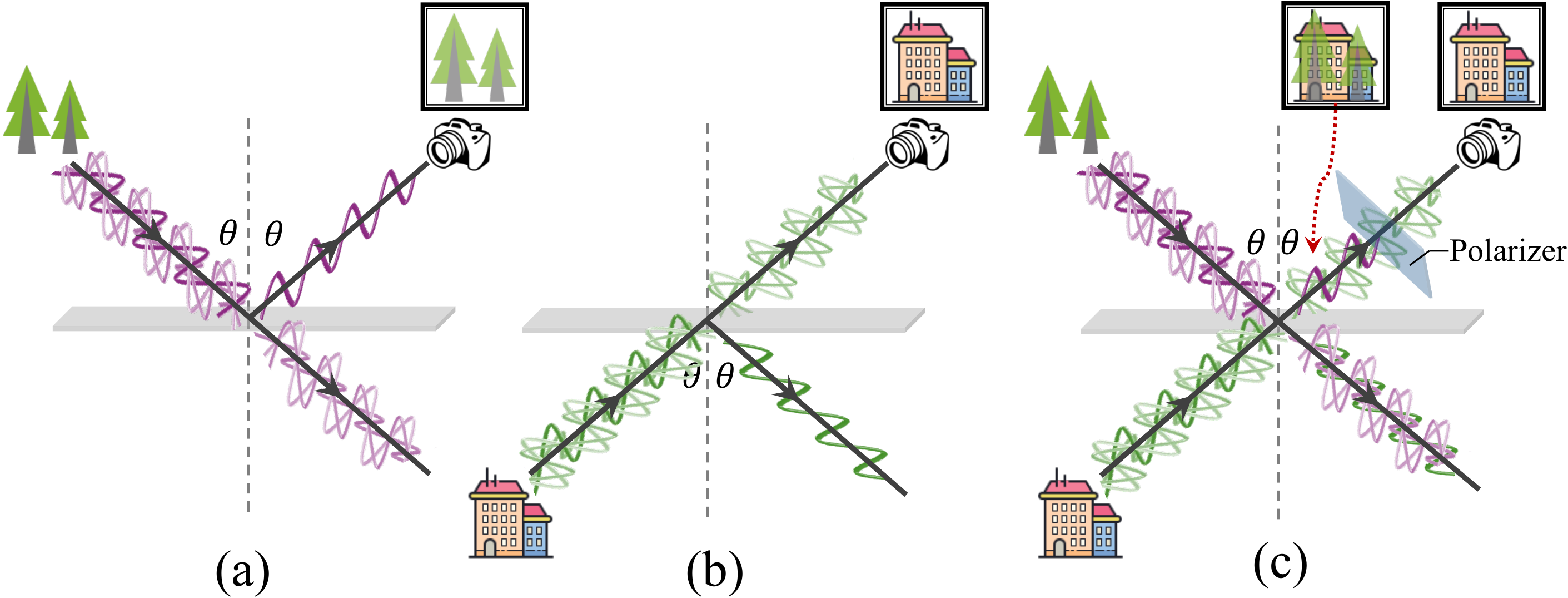}
	\caption{(a) \& (b) A semi-reflector transforms unpolarized light into polarized light upon reflection and refraction, which is undetectable by standard RGB cameras but can be leveraged by polarization cameras for reflection-suppression tasks. (c) At the Brewster angle~\cite{born2013principles}, a polarizer minimizes reflections.
}
	\label{fig:fig2}%
	\vspace{-0.7cm}
\end{figure}

Polarization image sensors are becoming mainstream, allowing users to easily capture polarization images from a single shot in real-time~\cite{sony_polarsens}.
However, existing methods~\cite{li2014single,zhu2024revisiting,hu2021trash,hu2023single} typically rely on intensity-based cues, such as pixel brightness and color gradients, to distinguish transmitted and reflected layers. These methods face challenges because reflection removal is a highly ill-posed inverse problem~\cite{li2020single} that recovers two unknown layers (reflection and transmission) from a single observation. Polarization provides valuable physics-based cues~\cite{nayar1997separation,schechner1999polarization} to alleviate ill-posed issues (Fig.~\ref{fig:fig1}), whereas transmitted light retains a distinct degree of polarization(Fig.~\ref{fig:fig2}b).
This difference provides crucial signals for separating the two layers. Notably, at the Brewster angle (Fig.\ref{fig:fig2}c), reflected light is fully polarized, enabling effective reflection removal~\cite{born2013principles}.


Despite significant advances of polarization images, a key challenge in polarization-based reflection removal is the lack of large-scale, high-quality datasets. Existing polarized reflection removal datasets~\cite{wieschollek2018separating,lei2020polarized} are limited in size and diversity, relying on small ($<$1000) or synthetic samples that fail to capture the complexity of real-world lighting conditions, materials, and scenes. 
Moreover, they typically exclude color information, reducing their applicability in real-world reflection removal tasks. Thus, there is a pressing need for a large-scale, comprehensive dataset that includes both RGB and polarization images, captured in diverse real-world environments, to advance polarization-based reflection removal.

To bridge this gap, we introduce PolaRGB, a novel dataset specifically collected for polarization-based reflection removal. As shown in Table~\ref{tab:tab1}, PolaRGB contains 6,500 high-quality, well-aligned RGB-polarization image pairs, 8$\times$ larger than previous dataset~\cite{lei2020polarized}. Our dataset covers a diverse range of scenes, lighting conditions, and exposure settings, and is captured using off-the-shelf commercial cameras with polarized color patterns to ensure real-world applicability.
PolaRGB provides both mixed images and ground-truth transmission layers, enabling accurate reflection separation and significantly enhancing the effectiveness of reflection removal across real-world scenarios.

Moreover, extracting reflection-free information from polarization data is challenging~\cite{wieschollek2018separating,lei2020polarized,lyu2019reflection} due to the randomness of shooting angles, scene variations, and changing lighting conditions. To address this issue, we leverage the powerful generative capabilities of the diffusion model~\cite{ho2020denoising, chen2024hierarchical,guan2024diffusion} to generate reflection-free cues. The diffusion model extracts and refines reflection-free priors from polarization images, effectively guiding the reflection removal and yielding precise and robust reflection-free results.

Specifically, PolarFree consists of two steps: a prior-generation step and a reflection removal step. First, in the prior-generation step, we use a diffusion model to genarate reflection-free prior based on the polarization and RGB inputs. This strategy not only guides accurate reflection isolation in RGB images but also recovers background details, which previous methods~\cite{zhu2024revisiting,kim2020single,hu2023single} may miss. Next, the reflection removal step leverages the prior to effectively remove reflections, ensuring accurate transmission restoration. Additionally, we introduce a phase-based loss function in the frequency domain to mitigate color discrepancies caused by semi-reflections, guiding the network to focus on reflection removal rather than color adjustment. These components enable PolarFree to achieve robust reflection suppression while preserving the clarity and integrity of the transmission across diverse real-world scenes.

Extensive experiments on the PolaRGB dataset demonstrate the effectiveness of PolarFree, which outperforms existing methods~\cite{zhu2024revisiting,kim2020single,hu2023single} 
$\sim$ 2dB in terms of PSNR. Additionally, we performed real-world testing in environments commonly affected by reflections such as museums and galleries, demonstrating the method's effectiveness in real-world scenarios. Our approach improves image clarity and preserves details better than previous methods, paving the way for more practical, reflection-robust applications. 

\begin{table}[t!]
\centering
\footnotesize
  \renewcommand\arraystretch{1.3}
\caption{
Comparisons between PolaRGB and existing reflection removal datasets.}\label{tab:tab1}
\vspace{-0.3cm}

\resizebox{1.0\linewidth}{!}{

\begin{tabular}{ccccccc}
\hline
\textbf{Dataset} & \textbf{Polarization} & \textbf{RGB} & \textbf{RAW} & \textbf{Data size} & \textbf{Resolution} \\ \hline
SIR$^2$~\cite{wan2019corrn}              & No                    & Yes          & No           & 500                   & 540 $\times$ 400    \\ 
RRW~\cite{zhu2024revisiting}              & No                    & Yes          & No           & 14,952                 & 2580 $\times$ 1460  \\ 
ReflectNet~\cite{wieschollek2018separating}   & Yes (Syn.)            & Yes           & No           & -                    & 1500 $\times$ 1000                   \\ 
Lei \emph{et al.}~\cite{lei2020polarized}            & Yes (Real)            & No           & Yes          & 807                   & 1224 $\times$ 1024                   \\ \hline
PolaRGB (Ours)             & Yes (Real)            & Yes          & Yes          & 6,500                 & 1224 $\times$ 1024  \\ \hline
\end{tabular}}
\vspace{-0.6cm}
\end{table}

\vspace{-0.4cm}
\section{Related Work}
\vspace{-0.2cm}

\subsection{Reflection Removal}
\vspace{-0.2cm}
Reflection removal~\cite{tan2005separating,schechner2000polarization,agrawal2005removing} has a longstanding history in computer vision, aiming to enhance the occluded objects caused by semi-reflectors like glass and windows. Early approaches~\cite{levin2007user, li2014single} typically relied on handcrafted priors, such as smoothness assumptions and gradient sparsity, to differentiate the reflection layer from the underlying scene. Some methods incorporated ghosting~\cite{shih2015reflection}, flash cues~\cite{lei2021robust,lei2023robust} and edge consistency \cite{han2017reflection}, while others improved accuracy by leveraging multi-image setups\cite{li2020improved, niklaus2021learned}, utilizing temporal variation~\cite{nandoriya2017video} and parallax~\cite{niklaus2021learned} to separate reflections from background layers. However, these methods often require strict assumptions and fail in real-world scenes with complex textures or lighting~\cite{wei2019single,lei2023robust}.

With the rise of deep learning~\cite{zhang2018single,wan2018crrn,zhu2024revisiting,yao2024neural,gou2020clearer}, reflection removal methods advanced considerably. Techniques based on neural networks have shown promise by learning to separate reflection and transmission layers directly from data~\cite{zhang2018single, wan2018crrn,wan2019corrn,hu2021trash,wang2023personalized,kim2020single}. However, without essential physical insights, these methods act as Bayesian regressions toward a learned average, often lead to suboptimal separation in complex environments with varying lighting, textures, and reflection intensities \cite{zhang2018single, wan2018crrn}. Recent methods~\cite{hongdiffer,zhong2024language} utilize language to interactively remove the reflection, which require additional efforts, limiting their practicality.

\begin{figure*}[!t]
    \centering
    \includegraphics[width=1\linewidth]{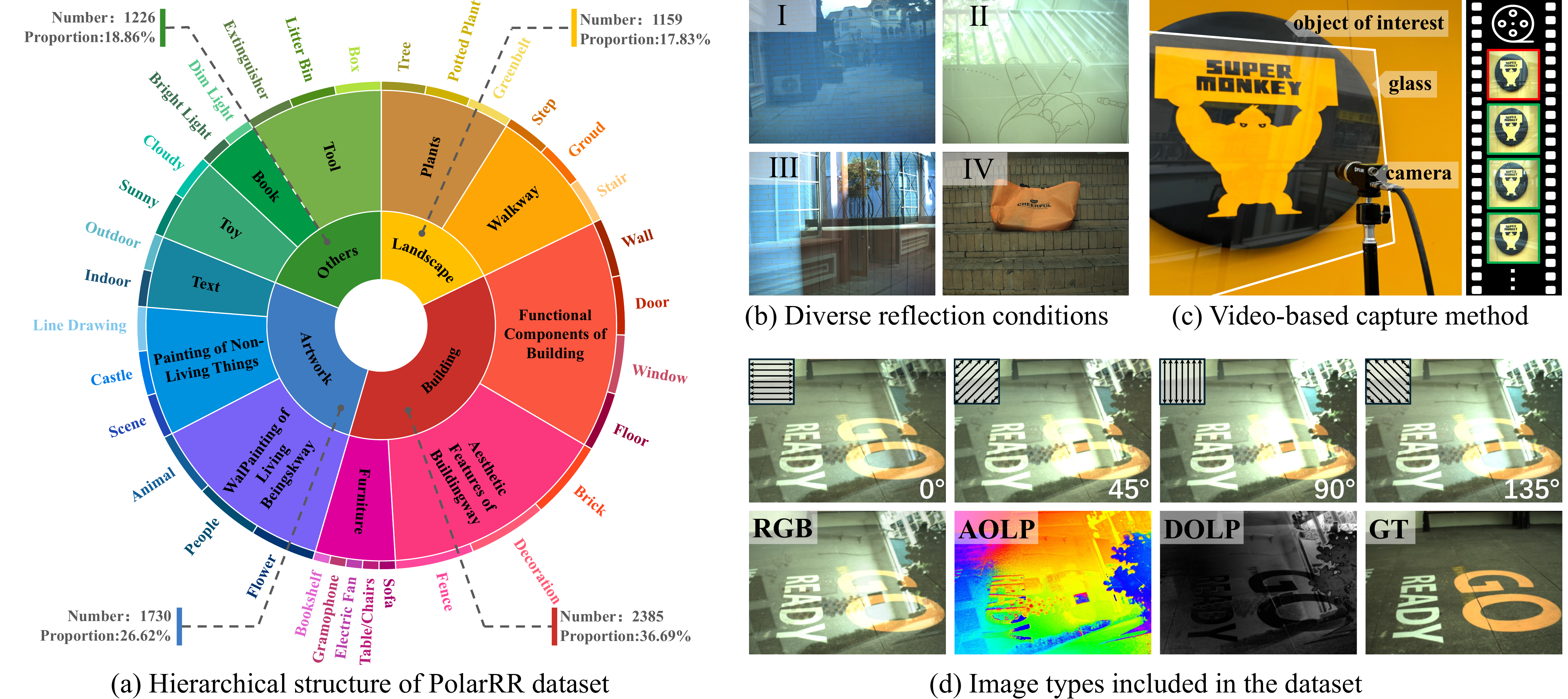}
        \vspace{-0.65cm}
    \caption{
    Overview of the PolaRGB dataset. (a) Hierarchical structure of scenes is shown in the ring, with legends indicating sample counts and subset types. (b) Typical scenes illustrating varied reflection conditions: I. smooth blending of reflection and refraction, II. abrupt reflection with mixed components, III. reflection dominant over transmission, and IV. minimal or no reflection. (c) Video-based capture method (details in Sec.~\ref{sec:PolaRGB Dataset}). (d) We provide polarized images at angles $\phi$ = 0$^\circ$, 45$^\circ$, 90$^\circ$, and 135$^\circ$, along with derived AoLP, DoLP, and a well-aligned unpolarized image. The dataset also includes ground truth transmission and estimated reflections, all available in both raw and RGB formats.}
    \label{fig:fig3}
    \vspace{-0.7cm}

\end{figure*}

\vspace{-0.2cm}
\subsection{Polarization Reflection Removal}
\vspace{-0.2cm}
Polarization is an effective way to remove reflections due to the physics principle: the polarization of light behaves differently in the reflection and transmission layers~\cite{nayar1997separation,schechner2000polarization,farid1999separating,li2023polarimetric,jeon2024spectral}, allowing separation, espicially at the Brewster angle~\cite{nayar1997separation,farid1999separating}. However, capturing images exactly at the Brewster angle, as shown in Fig.~\ref{fig:fig2}c, is challenging. Therefore, in practice, methods typically rely on images taken at multiple polarization angles (\emph{e.g.}, 0$^\circ$, 45$^\circ$, 90$^\circ$, and 135$^\circ$) to capture polarization information~\cite{nayar1997separation,schechner2000polarization,farid1999separating}.

Early polarized reflection removal methods~\cite{nayar1997separation,schechner2000polarization,farid1999separating} employ mathematical techniques like Pricinple Component Analysis (PCA) to separate reflection layers, which are effective when the transmission and reflection layers have significant content differences. Subsequently, Kong \emph{et al.}~\cite{kong2013physically} propose a multiscale scheme that automatically identifies the optimal separation of the reflection and background layers. Another advancements
~\cite{wieschollek2018separating} introduce a neural network-based approach for polarization-guided reflection removal with synthetic dataset. Lyu \emph{et al.}~\cite{lyu2019reflection} utilize a pair of unpolarized and polarized images but lack of realistic polarization patterns. Lei \emph{et al.}~\cite{lei2020polarized} further contribute by collecting a dataset of polarized images for reflection removal. However, this dataset only includes limited scene variations from pure polarization scenes, which limits its application scenario. In contrast, we present the first large-scale dataset captured for polarization-based RGB reflection removal, consisting of 6,500 well-aligned transmission-reflection pairs. We also propose a polarization-based reflection removal network based on diffusion models.

\begin{figure}[!t]
    \centering
    \includegraphics[width=1\linewidth]{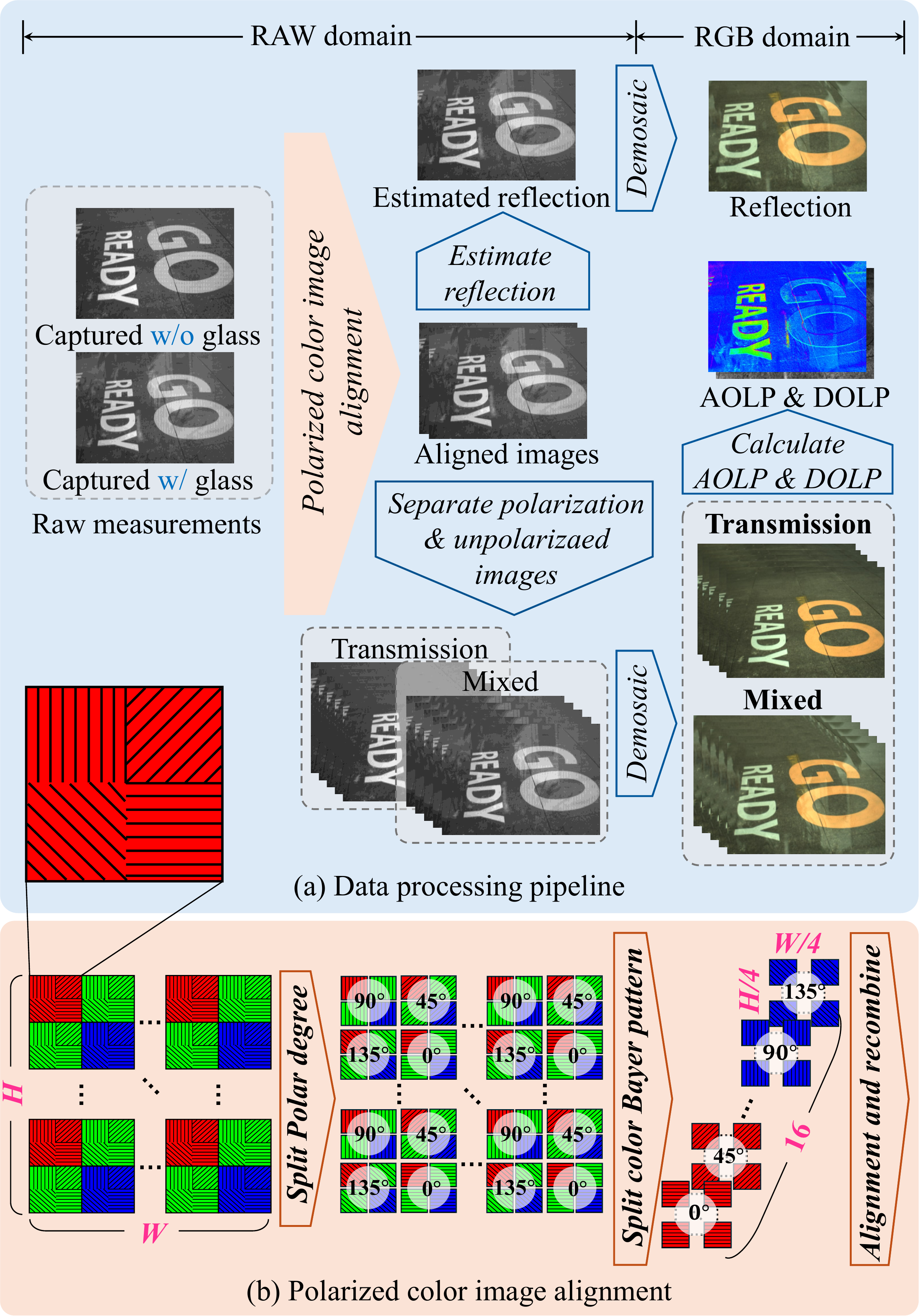}
    \vspace{-0.7cm}
    \caption{
     Data processing pipeline for obtaining aligned mixed and transmission images, and polarized images.}~\label{fig:fig3-5}
    \label{fig:warp_results}
    \vspace{-1.0cm}
\end{figure}

\begin{figure*}[!t]
    \centering
    \includegraphics[width=1\linewidth]{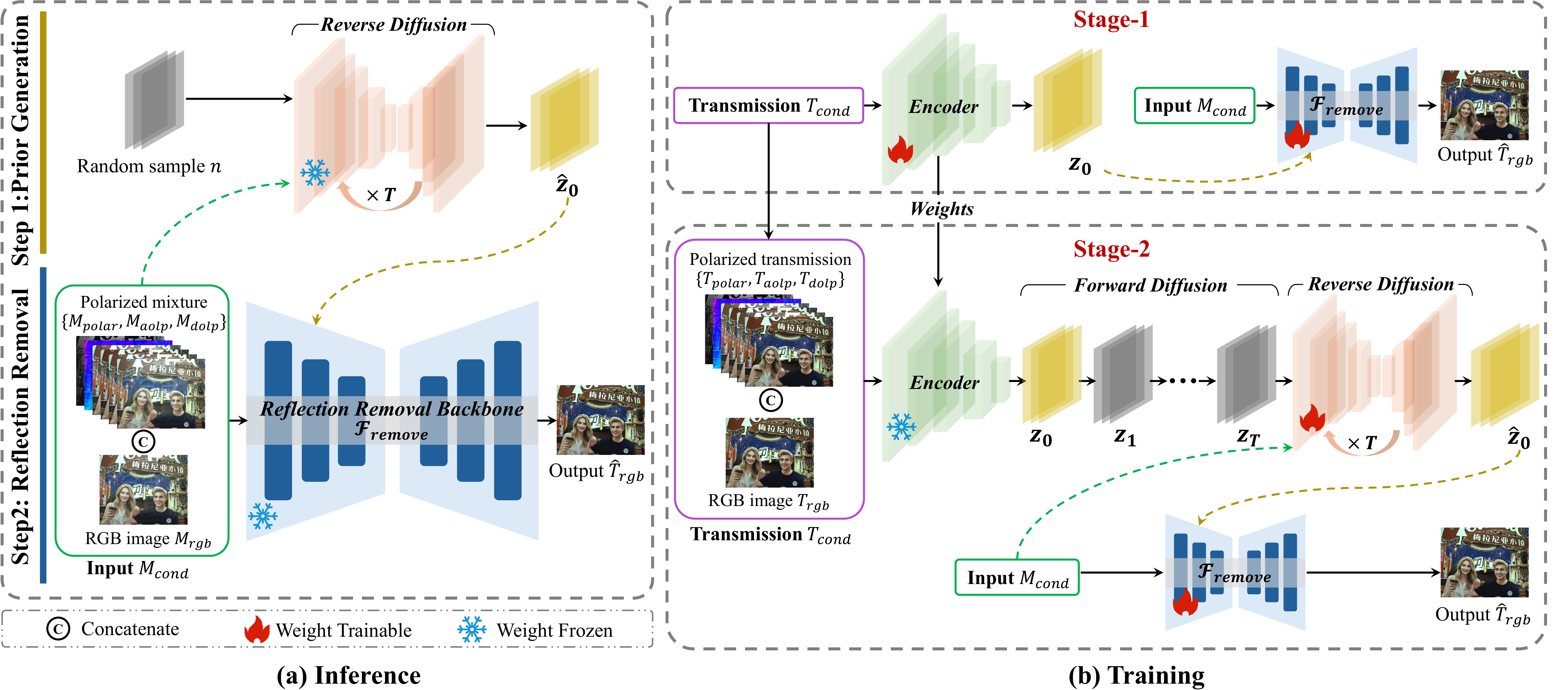}
        \vspace{-0.65cm}
    \caption{Pipeline of PolarFree. (a) During inference, PolarFree leverages polarized and RGB images as inputs, which are feeds into a conditional diffusion model to generate the prior $\hat{z}_0$. The generated prior, along with the inputs, is then passed to the reflection removal backbone $\mathcal{F}_{\text{remove}}$ to remove reflections. (b) PolarFree is trained in two stages. (1) A prior encoder extracts a reflection-free prior $z_0$ from clean transmission images and polarization cues, which serves as the supervision for the conditional diffusion model in stage two. (2) The conditional diffusion model is trained to progressively denoise noisy images, supervised by the prior from stage one, ensuring robust reflection separation.}~\label{fig:fig5}
 \vspace{-0.9cm}
\end{figure*}

\vspace{-0.4cm}
\section{PolaRGB Dataset}
\label{sec:PolaRGB Dataset}
\vspace{-0.1cm}

\subsection{Analysis}
\vspace{-0.1cm}
Although a few polarization-based reflection removal datasets exist~\cite{wieschollek2018separating,lei2020polarized}, they face major limitations: (1) they are often synthetically generated~\cite{wieschollek2018separating,lyu2019reflection}, which may not generalize well to real polarization, as perfectly simulating polarization phenomena is very hard; (2) they are typically limited in size and diversity ($<$1,000 samples), with limited scene variations, which constrains model robustness and generalizability~\cite{wieschollek2018separating,lei2020polarized}; and (3) they lack RGB data~\cite{wieschollek2018separating,lei2020polarized}, reducing the practical value of these datasets.

Our polarized-based reflection removal dataset, PolaRGB, addresses all these limitations with the following key features, as shown in Fig.~\ref{fig:fig3}. It has following advantages: (1) It contains real-captured images with perfect spatial alignment. We achieve pixel-level alignment through a three-step process: careful capture setup, manual filtering, and homography transformations, all in raw space to avoid demosaicing artifacts (see Fig.~\ref{fig:fig3-5}b and more details in supplementary). (2) It contains large-scale diverse scenes. The collected dataset covers a wide range of indoor and outdoor settings with four distinct reflection types, including smooth, sharp, high-brightness, and subtle reflections (see Fig.~\ref{fig:fig3}b). (3) It has extensive data modalities. Our dataset offers paired mixed and transmission images with both polarization and RGB captures, precisely aligned and available in both raw and RGB formats (see Fig.~\ref{fig:fig3}d).

\vspace{-0.2cm}
\subsection{Processing Pipeline}\label{sec:Processing pipeline}
\vspace{-0.2cm}
We leverage an efficient video-based capture flow for dataset collection, as shown in Fig.~\ref{fig:fig3}c. First, we capture a transmission-only image $T_{raw}$ as the ground truth. Then, we place a semi-reflective glass plate in front of the scene and rotate it continuously to capture mixed images $M_{raw}$. A division-of-focal-plane polarization camera with color Bayer pattern is used to simultaneously capture both polarized and RGB information, as shown in Fig.~\ref{fig:fig3-5}.

After obtaining the raw mixed image $M_{{raw}}$ and the transmission image $T_{{raw}}$, we process the images sequentially in the raw and RGB domains as shown in Fig.~\ref{fig:fig3-5}a. In the raw domain, we first align $M_{{raw}}$ and $T_{{raw}}$ to correct spatial misalignment caused by light refraction. This is done by separating the images into different polarization angles and color channels, then applying affine transformation matrices to each of the channels to avoid aliasing from direct alignment, as shown in Fig.~\ref{fig:fig3-5}b. Next, we perform polarization separation on the aligned images to obtain four polarization images ($0^\circ$, $45^\circ$, $90^\circ$, and $135^\circ$), and use Malus's law~\cite{schechner1999polarization,nayar1997separation} to sum them, producing the unpolarized image of the scene. Finally, we estimate the reflection image by searching for the optimal blending coefficient $\alpha_r$ and $\alpha_t$ in edge space, as illustrated in Eq.~\ref{eq:eq1}. For detailed steps and proof, please refer to the supplementary material.

Next, we apply demosaicking to the processed raw images to obtain the mixed, transmission, and reflection images, each consisting of four polarization images and the unpolarized RGB image. Additionally, we compute the Stokes parameters (see Sec.~\ref{sec:Problem Formulation}) to estimate the Angle of Linear Polarization (AOLP) and Degree of Linear Polarization (DOLP), which are then used in reflection removal.

\vspace{-0.3cm}
\section{Method}\label{sec:method}
\vspace{-0.1cm}

\subsection{Problem Formulation}
\label{sec:Problem Formulation}
\vspace{-0.1cm}

\paragraph{Object.} Given an RGB image and its corresponding polarization images, we aim to recover the transmission layer of the RGB image by utilizing the distinct polarization characteristics. Our measurement consists of spatially-aligned RGB and polarization images, where the polarization images captured at four distinct angles (0$^\circ$, 45$^\circ$, 90$^\circ$, and 135$^\circ$). These four angles provide a comprehensive polarization measurement~\cite{schechner2000polarization}, which allows us to compute polarization features crucial for separating layers.

\vspace{-0.3cm}
\paragraph{Polarization Prelimiary.}
For semi-reflective surfaces that produce mixed images, the observed intensity $I^{\phi}_M(x)$ can be decomposed into reflected intensity $I^{\phi}_R(x)$ and transmitted intensity $I^{\phi}_T(x)$ as
\begin{equation}
\abovedisplayskip=1pt  
\abovedisplayshortskip=3pt
I_M^{\phi}(x) = \alpha_R I_R^{\phi}(x) + \alpha_T I_T^{\phi}(x).
\belowdisplayskip=3pt  
\belowdisplayshortskip=1pt
\end{equation}
If a linear polarizer is placed in front of the camera at an angle $\phi$, the captured intensity $I^\phi(x)$ can be represented as
\begin{equation}
\abovedisplayshortskip=1pt
I_M^{\phi}(x) = \alpha(\theta; \phi; \phi_\perp)  I_R^{\phi}(x) + (1 - \alpha(\theta; \phi; \phi_\parallel))  I_T^{\phi}(x),
\belowdisplayshortskip=1pt
\end{equation}
where $\alpha(\theta; \phi; \phi_\perp)$ and $(1 - \alpha(\theta; \phi; \phi_\parallel))$ are polarization-dependent mixing coefficients based on the angle of incidence $\theta$, $\phi_\perp$ and $\phi_\parallel$ are the canonical polarization directions for the reflection and transmission, respectively. Directly solving for these parameters is challenging because $\alpha(\cdot)$, $I^{\phi}_R (x)$, and $I^{\phi}_R(x)$ are all unknown~\cite{nayar1997separation,wieschollek2018separating}. Additionally, relationship between the observed intensity and the underlying reflected and transmitted components is highly nonlinear and influenced by various factors~\cite{wieschollek2018separating}. To address this, we utilize Stokes parameters~\cite{lei2020polarized} that provide a efficient way to represent and analyze polarized light, enabling more robust separation of reflection and transmission.

\vspace{-0.3cm}
\paragraph{Stokes Parameters.} To capture the polarization effects in the scene, we use Stokes parameters $[S_0, S_1, S_2]$, which can be derived from intensity measurements at specific polarization angles (0$^\circ$, 45$^\circ$, 90$^\circ$, and 135$^\circ$):
\begin{equation}
\abovedisplayskip=1pt  
\abovedisplayshortskip=3pt
\begin{aligned}
S_0 &= (I_{0^\circ}+I_{45^\circ}+I_{90^\circ}+I_{135^\circ})/2, \\
S_1 &=  I_{0^\circ} - I_{90^\circ}, 
S_2 = I_{45^\circ} - I_{135^\circ}.
\end{aligned}
\belowdisplayskip=3pt  
\belowdisplayshortskip=1pt
\end{equation}
Here, $S_0$ represents the total intensity of light, and $S_1$ and $S_2$ provide information about the linear polarization state of the light based on intensity differences at these key angles. 

Using the Stokes parameters, we compute the Degree of Linear Polarization (DOLP) and Angle of Linear Polarization (AOLP) as follows~\cite{born2013principles}
\begin{equation}
\abovedisplayskip=1pt  
\abovedisplayshortskip=3pt
\begin{aligned}
{DOLP}(x) =& {\sqrt{S_1(x)^2 + S_2(x)^2}}/{S_0(x)}, \\
\quad {AOLP}(x) =& \frac{1}{2} \text{atan2} \left( {S_2(x)}/{S_1(x)} \right).
\end{aligned}
\belowdisplayskip=3pt  
\belowdisplayshortskip=1pt
\end{equation}
In this formulation, DOLP describes the proportion of polarized light relative to total intensity, which helps indicate the degree of reflection in the scene. AOLP, on the other hand, reveals the orientation of the polarized light, allowing us to distinguish between reflection and transmission components more effectively, as shown in~\cref{fig:fig3}d. By leveraging intensity measurements at 0$^\circ$, 45$^\circ$, 90$^\circ$, and 135$^\circ$, these parameters provide valuable cues for separating mixed reflection and transmission layers in semi-reflective scenes.

\vspace{-0.2cm}
\subsection{PolarFree Network}
\vspace{-0.2cm}
To achieve high-quality reflection removal with effective utilization of polarization information, we introduce PolarFree, a dedicately designed two-step network, where each step addresses a distinct aspect of the reflection removal challenge. Inspired by~\cite{chen2024hierarchical}, we leverage difusion model to generate the prior for reflection removal. As shown in Fig.~\ref{fig:fig5}, during inference, the first step utilizes a conditional diffusion model to extract reflection-free priors, effectively isolating essential details from polarization data. The second step integrates these priors with RGB inputs, guiding the network to accurately separate reflections and enhance clarity, even in complex, real-world environments. 

\vspace{-0.3cm}
\paragraph{Prior Generation.} As shown in Fig.~\ref{fig:fig5}a, in the first step, we start with a randomly initialized noise $n$, which is gradually denoised through a conditional diffusion model $\mathcal{F}_\text{diff}$
\vspace{-0.3cm}
\begin{equation}
\hat{z}_0 = \mathcal{F}_\text{diff}(n| M_\text{cond}),   
\end{equation}
where $\hat{z}_0$ is the generated prior, $M_\text{cond}=\{M_\text{polar}, M_\text{aolp}, M_\text{dolp}, M_\text{rgb}\}$ represents the mixed images, and $n$ is the initial noise. Through an iterative denoising process, the diffusion model progressively refines a noise image, conditioned on the polarization and RGB data, to generate a reflection prior.

The denoising process follows the denoising diffusion probabilistic model (DDPM) framework~\cite{ho2020denoising} and employs a U-Net architecture to predict noise. 
At each timestep $t$, the U-Net receives a noisy intermediate $z_t$ and outputs a noise estimate $\epsilon_\theta(z_t, t)$, predicting the noise added at that timestep. This process can be formulated as
\begin{equation}~\label{eq:eq7}
\abovedisplayskip=1pt  
\abovedisplayshortskip=3pt
z_{t-1} = \frac{1}{\sqrt{\alpha_t}} \left( z_t - \frac{\beta_t}{\sqrt{1 - \bar{\alpha}_t}} \epsilon_\theta(z_t,M_\text{cond}, t) \right) + \sigma_t z,
\belowdisplayskip=3pt  
\belowdisplayshortskip=1pt
\end{equation}
where $\alpha_t$ and $\beta_t$ control the noise schedule across the timesteps, $\sigma_t$ is the standard deviation, and $z$ is noise sampled from a standard Gaussian distribution. $\bar{\alpha}_t$ is the cumulative product of $\alpha_t$, indicating noise level to timestep $t$.

In this way, the U-Net gradually removes the noise in each step, conditioning on the polarization-based measurement input $M_\text{cond}$. This process continues until noise is fully removed, yielding a sample $\hat{z}_0$ that represents the reflection-free prior distribution extracted from the conditioned inputs.

\vspace{-0.6cm}
\paragraph{Reflection Removal.} As shown in Fig.~\ref{fig:fig5}a, once the prior $\hat{z}_0$ has been obtained, the second step of PolarFree is seperating the transmission and reflection layers. By leveraging the polarization cues provided by $M_\text{polar}$, $M_\text{aolp}$, and $M_\text{dolp}$, the model $\mathcal{F}_\text{remove}$ separates the transmission features from the reflection ones, which can be represented as 
\begin{equation}
\abovedisplayskip=1pt  
\abovedisplayshortskip=3pt
\hat{T}_{rgb} = \mathcal{F}_\text{remove}(\hat{z}_0, M_\text{cond}),
\belowdisplayskip=3pt  
\belowdisplayshortskip=1pt
\end{equation}
where $\mathcal{F}_\text{remove}$ is a reflection removal neural network. This step ensures that the final output contains distinct transmission and reflection components, facilitating high-quality image reconstruction under challenging reflective scenarios.

\begin{figure}[t]
    \centering
    \includegraphics[width=0.98\linewidth]{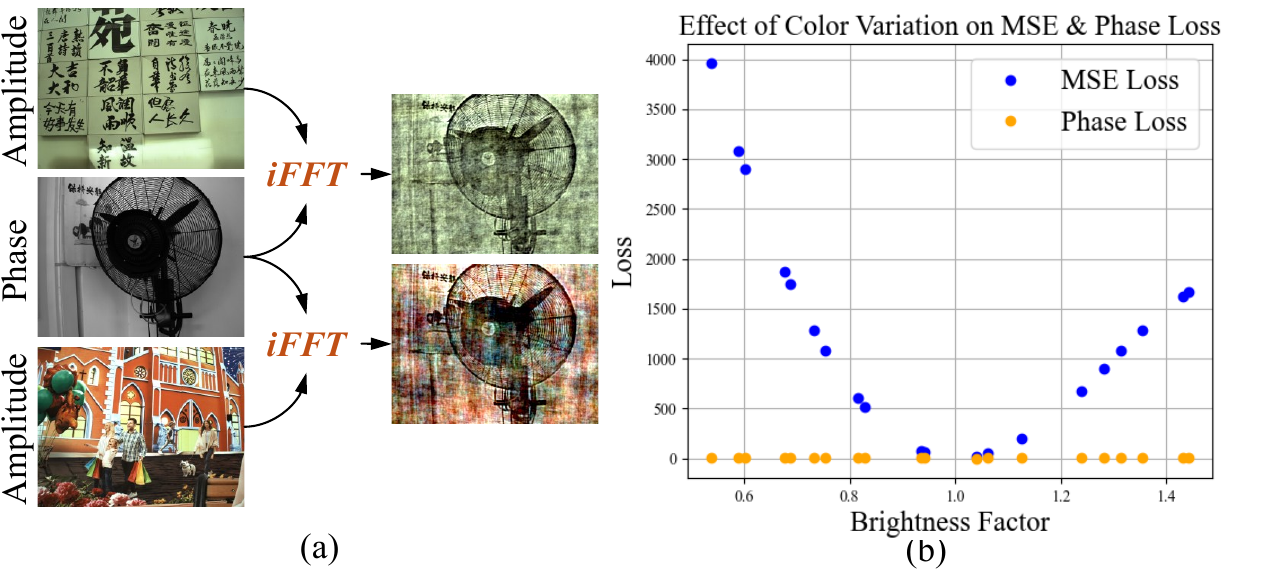}
    \vspace{-0.4cm}
    \caption{ (a) Phase information preserves shape and texture details, while color primarily affects the amplitude. (b) We apply two types of random color perturbations to the image and compute the perturbation errors. Phase-based loss is less sensitive to color changes.
    }
\label{fig:fig6}
\vspace{-0.7cm}
\end{figure}

\begin{figure*}[!t]
    \centering
    \includegraphics[width=1\linewidth]{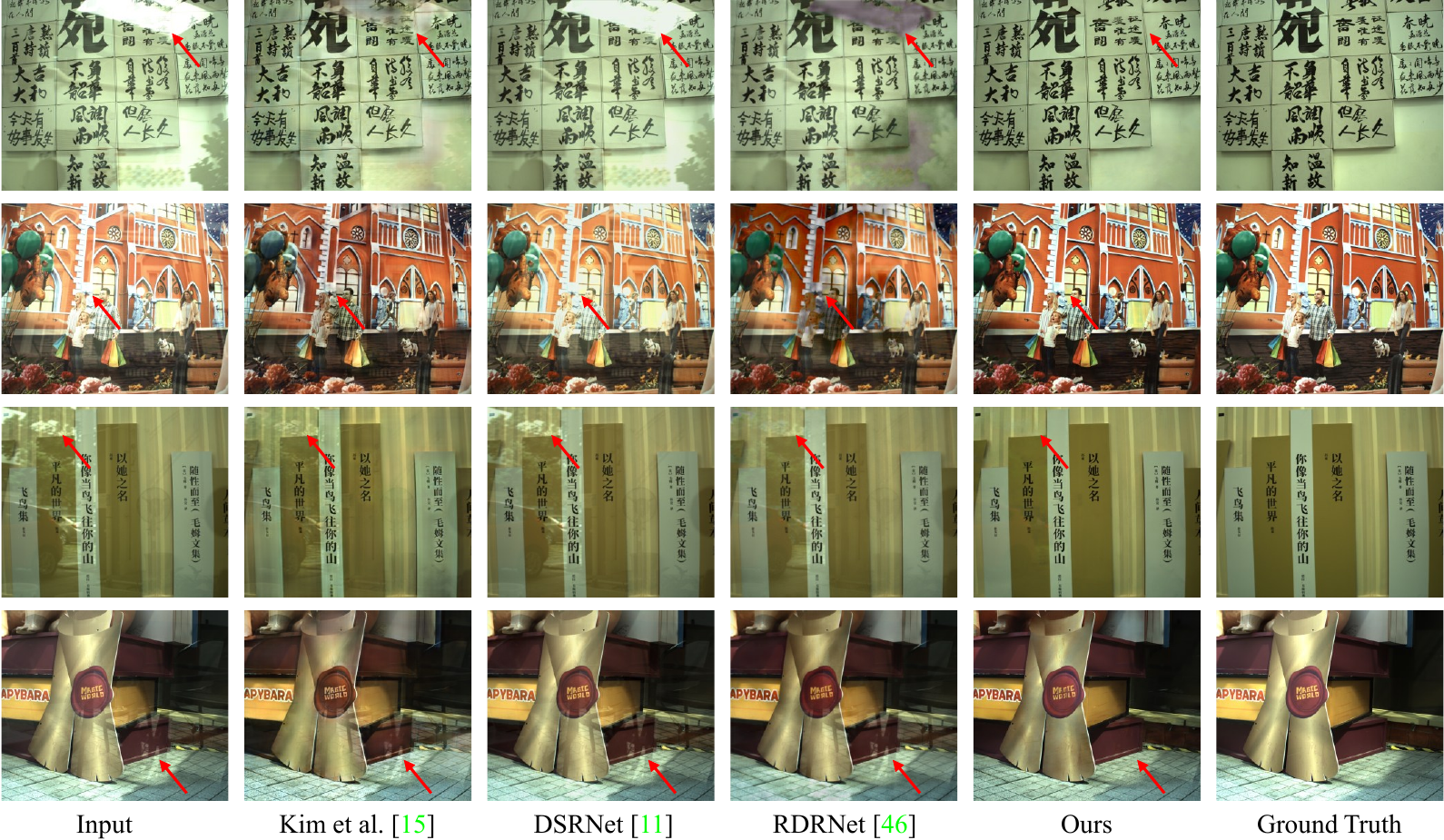}
    \vspace{-0.7cm}
    \caption{Qualitative comparisons on the PolaRGB dataset.}\label{fig:fig7}
\vspace{-0.6cm}
\end{figure*}

\vspace{-0.2cm}
\subsection{Training}
\vspace{-0.1cm}
To train our PolarFree, particularly to address the challenge of the diffusion model lacking a suitable ground truth, we adopt a two-stage training strategy, as illustrated in Fig.~\ref{fig:fig5}b. This strategy consists of two sequential objectives: \textit{learning to extract a reflection-free prior} and \textit{learning to generate a reflection-free prior}.

\vspace{-0.5cm}
\paragraph{First Stage.} In the first stage, we train an encoder to extract a prior $z_0$ for reflection-free information, which will serve as the supervision signal for the diffusion model in the second stage. Specifically, we feed the ground truth transmission images $T_\text{cond}=\{T_\text{polar}, T_\text{aolp}, T_\text{dolp}, T_\text{rgb}\}$ into the encoder $\mathcal{E}$, obtaining $z = \mathcal{E}(M_\text{cond}, T_{\text{rgb}})$. This $z$ contains reflection-free information enriched with polarization-related cues. Subsequently, $z_0$ conditions the reflection removal backbone $\mathcal{F}_\text{remove}$ to predict the clean transmission image $\hat{T}_\text{rgb} = \mathcal{F}_\text{remove}(z, M_\text{cond} )$.

\vspace{-0.5cm}
\paragraph{Second Stage.} In the second stage, we train the diffusion model $\mathcal{F}_\text{diff}$ to generate a reflection-free prior $z_0$ from noisy input images, and finetune the reflection removal backbone $F_\text{remove}$. The key challenge here is that the diffusion model lacks of direct supervision. Therefore, we leverage the prior $z_0$ extracted in the first stage as the supervision signal to guide the model.

We start with the extracted reflection-free prior $z_0$ and add noise over multiple timesteps. This process transforms the ``ground-truth'' prior $z_0$ into noisy versions $z_t$ at each timestep $t$, which can be expressed as
\begin{equation}~\label{eq:eq9}
\abovedisplayskip=1pt  
\abovedisplayshortskip=3pt
q(z_t | z_0) = \mathcal{N}(z_t; \sqrt{1 - \beta_t} z_0, \beta_t \mathbf{I}),
    \belowdisplayskip=3pt  
\belowdisplayshortskip=1pt
\end{equation}
where $\mathcal{N}$ represents a Gaussian distribution, $\beta_t$ is the noise schedule controlling the noise level added at each timestep, $\mathbf{I}$ represents the identity matrix.

In the reverse diffusion process, the model is trained to progressively remove the noise step by step to recover the clean prior $z_0$, which can be repersented as 
\begin{equation}
\abovedisplayskip=1pt  
\abovedisplayshortskip=3pt
p_\theta(z_{t-1} | z_t) = \mathcal{N}(z_{t-1}; \mu_\theta(z_t, t,M_\text{cond}), \sigma_t^2 \mathbf{I})),
    \belowdisplayskip=3pt  
\belowdisplayshortskip=1pt
\end{equation}
where $\mu_\theta(z_t, t, M_\text{cond})$ is the mean function of the current state $z_t$, timestep $t$, and a conditional input $M_\text{cond}$.

Throughout the reverse process, the model conditions on the polarization measurements $M_\text{cond}$, which provide essential information about the scene’s physical properties. These measurements help the model accurately generate reflection-free components during the denoising process.

After the reverse process completes, the model outputs a final reflection-free prior $\hat{z}_0$, which contains clean transmission information. This prior is used as the guidance signal for the reflection removal backbone $\mathcal{F}_\text{remove}$. The backbone takes this reflection-free prior and the polarization data to predict the clean transmission image $\hat{T}_{rgb}$.

\vspace{-0.2cm}
\subsection{Losses}
\vspace{-0.1cm}
\paragraph{Basic Loss.} To optimize PolarFree, we follow~\cite{zhu2024revisiting} to utilzie three basic losses: L1 loss, VGG perceptual loss~\cite{johnson2016perceptual}, and total variation (TV) loss~\cite{strong2003edge}.
The L1 loss minimizes the pixel-wise difference between the predicted transmission image $\hat{T}_{\text{rgb}}$ and the ground truth transmission image $T_{\text{rgb}}$ as $\mathcal{L}_{\text{1}} = \| \hat{T}_{\text{rgb}} - T_{\text{rgb}} \|_1$.
The VGG perceptual losscompares feature activations from a pre-trained VGG network~\cite{johnson2016perceptual} , weighted by $\lambda_l$: $\mathcal{L}_{\text{VGG}} = \sum_{l} \lambda_l \| \phi_l(\hat{T}_{\text{rgb}}) - \phi_l(T_{\text{rgb}}) \|_1$, where $\phi_l$ denotes the activations of the $l$-th VGG layer. We also use TV loss to constrain consistency via the gradient operator $\mathcal{L}_{\text{TV}} = \| \nabla \hat{T}_{\text{rgb}}- \nabla T_\text{rgb} \|_1$.

\vspace{-0.4cm}
\paragraph{Phase loss.} While basic losses help the network match color and intensity values to the ground truth, they struggle with color discrepancies caused by semi-reflective surfaces during dataset collection. These variations, due to reflection and transmission properties of materials, affect the color and intensity, leading to mismatches between the predicted and target images and hindering the model's ability to learn the correct transmission map.

To address this issue, we introduce a phase loss to focus on the structural information of the transmission, which is less sensitive to color variations. As shown in Fig.~\ref{fig:fig6}, phase information primarily captures the geometry and texture of the image, independent of color changes. The phase loss is formulated as
\begin{equation}
\abovedisplayskip=1pt  
\abovedisplayshortskip=3pt
    \mathcal{L}_{phase} = \| \angle(FFT(\hat{T})) - \angle(FFT((T_{rgb})) \|_1,
    \belowdisplayskip=3pt  
\belowdisplayshortskip=1pt
\end{equation}
where $FFT$ is the Fourier transform, and $\angle(\cdot)$ represents the phase angle of the Fourier coefficients.

\vspace{-0.4cm}
\paragraph{Diffusion Loss.} The diffusion loss follows the standard DDPM formulation~\cite{ho2020denoising}, where the model predicts the noise at each step $t$ and minimizes the difference between the predicted noise and the true noise as
\begin{equation}
\abovedisplayskip=1pt  
\abovedisplayshortskip=3pt
\mathcal{L}_{{diff}} = \mathbb{E}_{q(z_t | z_{t-1})} \left[ \|\epsilon_{\theta}(z_t, t) - \epsilon_{\text{true}}(z_t)\|_2^2 \right].
  \belowdisplayskip=3pt  
\belowdisplayshortskip=1pt
\end{equation}
Here, $\epsilon_{{true}}(z_t)$ is the added noise obtained by Eq.~\ref{eq:eq9}, and $\epsilon_{\theta}(z_t, t)$ is the predicted noise, conditioned on the polarization measurements and RGB input $M_\text{cond}$.

\vspace{-0.5cm}
\paragraph{Total Loss.}
The above losses are weighted summed to serve as the supervision for the first stage as
\begin{equation}
\abovedisplayskip=1pt  
\abovedisplayshortskip=3pt
  \mathcal{L}_{\text{stage1}} = \gamma_{{1}}  \mathcal{L}_{\text{1}} + \gamma_{{2}} \mathcal{L}_{\text{VGG}} + \gamma_{{3}} \mathcal{L}_{\text{TV}} +\gamma_{4} \mathcal{L}_{\text{phase}}.
  \belowdisplayskip=3pt  
\belowdisplayshortskip=1pt
\end{equation}
In the second stage, we use a combined loss function consisting of the diffusion loss and the reconstruction loss as
\vspace{-0.4cm}
\begin{equation}
\abovedisplayshortskip=3pt
\mathcal{L}_{\text{stage2}} = \gamma_5\mathcal{L}_\text{diff} + \gamma_6\mathcal{L}_{\text{recon}}.
\belowdisplayshortskip=1pt
\end{equation}
where $\mathcal{L}_\text{recon}$ has the same format as $\mathcal{L}_\text{stage1}$, and $\gamma_{(\cdot)}$ are the coefficients. This combined loss helps refine the reflection-free image while ensuring it aligns with the ground truth.

\vspace{-0.3cm}
\section{Experiments}
\vspace{-0.15cm}

\subsection{Implementation Details}
\vspace{-0.15cm}
We train and evaluate PolarFree on the PolaRGB dataset. The whole dataset consists of 67 scenes, with a total of 6,500 paired images. For each scene, we keep the background (transmission) and camera fixed, and varying the glass position to capture images with reflections. The 67 scenes are then randomly split into 56 training scenes and 11 testing scenes, containing 6,312 and 188 paired images, respectively. This division ensures no data leakage between the training and testing sets, with each set containing only a subset of all categories.


We also test PolarFree in real-world scenes like museums and galleries, where ground truth is unavailable. We implement PolarFree using PyTorch on a single NVIDIA RTX 4090 GPU. Training is conducted with a batch size of 2 and an AdamW optimizer with a learning rate of $2\times10^{-4}$, 30k iterations on the PolaRGB dataset for both stages.

We compare our method with recent advanced reflection removal methods, including Lei \emph{et al.}~\cite{lei2020polarized}, IBCLN~\cite{li2020single}, DSRNet~\cite{hu2023single}, YTMT~\cite{hu2021trash}, and RDRNet~\cite{zhu2024revisiting}. For a fair comparison, we modify the input settings to align with ours and use only the transmission layer for supervision. We have re-trained baseline methods on the PolaRGB dataset. The evaluation is conducted using both objective metrics (PSNR, SSIM), a perceptual metric (LPIPS), and a language-based non-reference metric (Q-Align~\cite{wu2023q}).

\vspace{-0.2cm}
\subsection{Results}
\vspace{-0.2cm}
\paragraph{Results on PolaRGB.}
We present the quantitative results in~\cref{tab:tab2}. It can be seen that, our method outperforms other methods across multiple metrics, demonstrating the effectiveness of PolarFree for high-quality reflection removal. Notably, for fairness, we modified the input layer of all baselines to accept polarization information. Visual results in Fig.~\ref{fig:fig7} show that PolarFree provides cleaner reflection removal with sharper edges and better color preservation. In contrast, previous methods often suffer from color distortions or imperfect reflection removal, especially in areas with complex reflections. Our method maintains high fidelity to the ground truth, especially in challenging regions with low light and subtle reflections.

\begin{figure}[!t]
    \centering
    \includegraphics[width=1\linewidth]{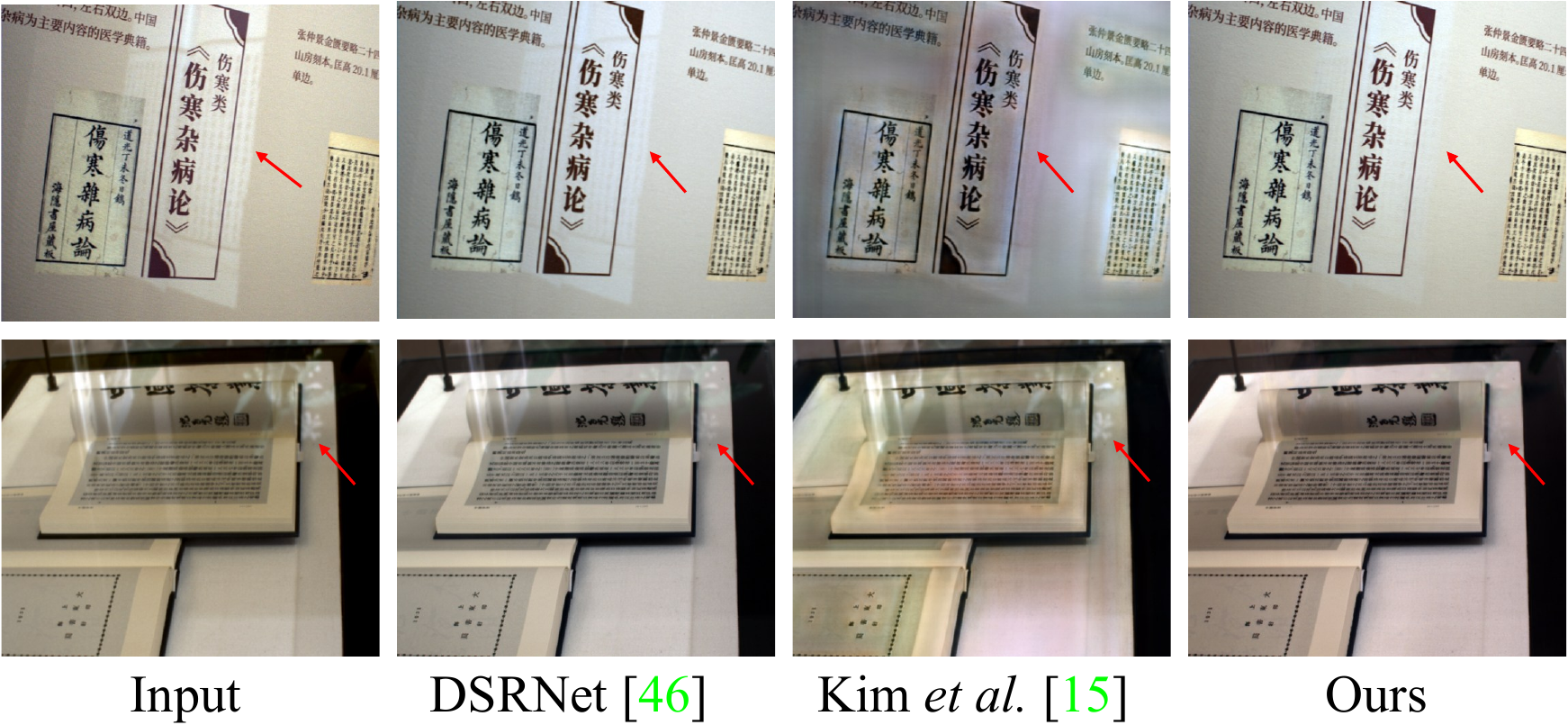}
    \vspace{-0.7cm}
    \caption{Visual comparison of reflection removal results in real-world museum scenes with glass display cases. 
    }
\label{fig:fig8}
\vspace{-0.3cm}
\end{figure}

\vspace{-0.4cm}
\paragraph{Real-captured without Ground Truth.}
To demonstrate the generalization ability to unseen and more complex reflection, We further evaluate our model on real-captured images at museum with complex reflections from glass enclosures. Here, we only provide qualitative results to visually assess the model’s capability. As shown in Fig.~\ref{fig:fig8}, our approach effectively reduces reflections while preserving fine details, despite the complex lighting and material variations often present in museum environments. This demonstrates the practical robustness of our method and its potential for real-world applications.

\vspace{-0.2cm}
\subsection{Ablation Study}
\vspace{-0.1cm}
We perform ablation studies to evaluate the effectiveness of key components in our method. More experimental results and analyses can be found in the supplementary.

\begin{table}[!t]
\centering
\footnotesize
\caption{
Quantitative comparison on the PolaRGB dataset, evaluated using objective metrics (PSNR, SSIM), perceptual metric (LPIPS), and the Language-based non-reference metric (Q-align~\cite{wu2023q}). Best results are in \textbf{bold}, second-best results are \underline{underlined}.
}\label{tab:tab2}
\vspace{-0.2cm}
\resizebox{1.0\linewidth}{!}{
\begin{tabular}{cccccccc}
    \toprule
    {Method} & {PSNR↑} & {SSIM↑} & {LPIPS↓} & {Q-align↑}\\
    \midrule
    Lei \emph{et al.}~\cite{lei2020polarized}  & 18.73 & 0.7962 &  0.3804 & 3.2109 \\
    Kim \emph{et al.}~\cite{kim2020single}  & \underline{20.67} & \underline{0.8399} & 0.2714 & \underline{3.7148} \\
    IBCLN~\cite{li2020single} & 19.73 & 0.8173  & \underline{0.2488} & 3.0938\\
    YTMT~\cite{hu2021trash}  & 16.86 &  0.7544 & 0.4489 & 3.0938 \\
    DSRNet~\cite{hu2023single}  & 16.84 & 0.7913 & {0.2828} & {3.6992}  \\
    RDRNet~\cite{zhu2024revisiting}  & 15.88 & 0.6964 & 0.5250 & 2.9531 \\
    \hline
    Ours  & \textbf{22.44} & \textbf{0.8681} & \textbf{0.1325} & \textbf{3.8867} \\

\bottomrule
\end{tabular}
}
\vspace{-0.6cm}
\end{table}

\begin{figure}[!t]
    \centering
    \includegraphics[width=1\linewidth]{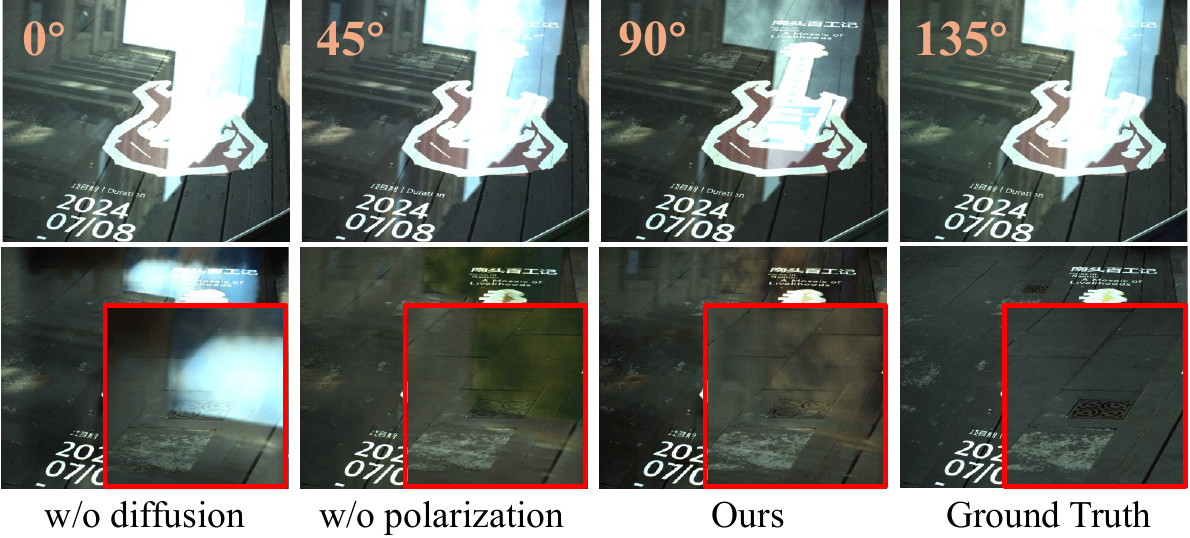}
        \vspace{-0.7cm}
    \caption{
Highly challenging reflection removal scenes with complex reflections and highlights. Top: Polarization images from different angles provide complementary information for effective reflection removal, note the third (90$^\circ$) image. Bottom: The diffusion module effectively handles highlights, while polarization better restores the color and details in such challenging scenarios.
       }~\label{fig:fig9}
        \vspace{-0.4cm}
\end{figure}

\vspace{-0.5cm}
\paragraph{Polarization Information. }
To assess the importance of polarization cues, we conduct an experiment where the polarization information (AOLP, DOLP, and polarization images) is removed, and the model is trained solely using RGB data. As shown in Table~\ref{tab:tab3} and Fig.~\ref{fig:fig9}, the absence of polarization significantly impairs the model’s ability to differentiate between reflection and transmission layers, leading to noticeably poorer reflection removal performance. These results demonstrate the critical role of polarization in accurate reflection removal.

\vspace{-0.4cm}
\paragraph{Diffusion Prior.} We also evaluate the effectiveness of diffusion prior by removing the conditional diffusion model. While without diffusion prior achieves reasonable results, as indicated by the comparison metrics in Table~\ref{tab:tab3}, it lacks fine-grained control and high-quality output offered by the diffusion model, especially in complex scenes (Fig.~\ref{fig:fig9}). The conditional diffusion process significantly improves separation accuracy and preserves finer transmission details.

\vspace{-0.4cm}
\paragraph{Phase-based Loss.} We explore the impact of our phase-based loss function. We remove this loss from the training procedure and observe the model’s performance without it. The results show a noticeable increase in color discrepancies and less accurate reflection removal, particularly in scenarios involving semi-reflective surfaces (Table~\ref{tab:tab3}). The phase-based loss function ensures the model focuses on reflection removal rather than compensating for color inconsistencies, leading to improved overall results.

\begin{table}[t]
\centering
\footnotesize
\vspace{-0.4cm}
\caption{Ablation study on PolarFree framework.}~\label{tab:tab3}

\vspace{-0.4cm}
\begin{tabular}{ccccc}
\toprule
{Method} & {PSNR↑} & {SSIM↑} & {LPIPS↓} & Q-align↑ \\
\midrule
w/o $\mathcal{L}_{phase}$ & {22.41} & 0.8622 & 0.1402 & 3.8524\\
w/o polarization & 21.56 & 0.8620 &  0.1483 & 3.7949\\
w/o diffusion prior & 20.21 & 0.8627 & 0.1552 &3.7949 \\
Ours & \textbf{22.44} & \textbf{0.8681} & \textbf{0.1325} & \textbf{3.8867} \\
\bottomrule
\end{tabular}
\vspace{-0.6cm}
\end{table}

\vspace{-0.3cm}
\section{Conclusion}\label{sec:Conclusion}
\vspace{-0.15cm}

In this work, we propose PolaRGB, a large-scale, comprehensive dataset for polarization-based reflection removal. With 6,500 well-aligned RGB-polarization image pairs, PolaRGB is 8$\times$ larger than existing polarization datasets and is the first to include both RGB and polarization images, uniquely captured across diverse real-world scenes and lighting conditions. Additionally, we present PolarFree, a novel reflection removal model that leverages the generative capabilities of diffusion models. This approach enables PolarFree to generate reflection-free cues, enhancing separation accuracy while preserving fine details in the transmission layer. A phase-based loss function is also introduced to further improve the model’s performance. Comprehensive experimetnal results on realistic scenarios demonstrate the effectiveness our method. We believe these contributions set a foundation for further advancements in polarization-based reflection removal, paving the way for more sophisticated applications in complex real-world environments.

{\small
\bibliographystyle{ieee_fullname}
\bibliography{egbib}

\begin{thebibliography}{10}\itemsep=-1pt

\bibitem{agrawal2005removing}
Amit Agrawal, Ramesh Raskar, Shree~K Nayar, and Yuanzhen Li.
\newblock Removing photography artifacts using gradient projection and flash-exposure sampling.
\newblock {\em ACM Transactions on Graphics}, 24(3):828--835, 2005.

\bibitem{born2013principles}
Max Born and Emil Wolf.
\newblock {\em Principles of optics: electromagnetic theory of propagation, interference and diffraction of light}.
\newblock Elsevier, 2013.

\bibitem{chen2024hierarchical}
Zheng Chen, Yulun Zhang, Ding Liu, Jinjin Gu, Linghe Kong, Xin Yuan, et~al.
\newblock Hierarchical integration diffusion model for realistic image deblurring.
\newblock {\em Advances in neural information processing systems}, 36, 2024.

\bibitem{farid1999separating}
Hany Farid and Edward~H Adelson.
\newblock Separating reflections and lighting using independent components analysis.
\newblock In {\em Proceedings. 1999 IEEE Computer Society Conference on Computer Vision and Pattern Recognition (Cat. No PR00149)}, volume~1, pages 262--267. IEEE, 1999.

\bibitem{gou2020clearer}
Yuanbiao Gou, Boyun Li, Zitao Liu, Songfan Yang, and Xi Peng.
\newblock Clearer: Multi-scale neural architecture search for image restoration.
\newblock {\em Advances in neural information processing systems}, 33:17129--17140, 2020.

\bibitem{guan2024diffusion}
Yuanshen Guan, Ruikang Xu, Mingde Yao, Ruisheng Gao, Lizhi Wang, and Zhiwei Xiong.
\newblock Diffusion-promoted hdr video reconstruction.
\newblock {\em arXiv preprint arXiv:2406.08204}, 2024.

\bibitem{han2017reflection}
Byeong-Ju Han and Jae-Young Sim.
\newblock Reflection removal using low-rank matrix completion.
\newblock In {\em Proceedings of the IEEE Conference on Computer Vision and Pattern Recognition}, pages 5438--5446, 2017.

\bibitem{ho2020denoising}
Jonathan Ho, Ajay Jain, and Pieter Abbeel.
\newblock Denoising diffusion probabilistic models.
\newblock {\em Advances in neural information processing systems}, 33:6840--6851, 2020.

\bibitem{hongdiffer}
Yuchen Hong, Haofeng Zhong, Shuchen Weng, Jinxiu Liang, and Boxin Shi.
\newblock L-differ: Single image reflection removal with language-based diffusion model.
\newblock In {\em Proceedings of the european conference on computer vision (ECCV)}, 2024.

\bibitem{hu2021trash}
Qiming Hu and Xiaojie Guo.
\newblock Trash or treasure? an interactive dual-stream strategy for single image reflection separation.
\newblock {\em Advances in Neural Information Processing Systems}, 34, 2021.

\bibitem{hu2023single}
Qiming Hu and Xiaojie Guo.
\newblock Single image reflection separation via component synergy.
\newblock In {\em Proceedings of the IEEE/CVF International Conference on Computer Vision}, pages 13138--13147, 2023.

\bibitem{huang2016dust}
Zhiyong Huang, Biao Xiong, Cao Tian, Jing Zhan, Xiang Fei, and Nazaraf Shah.
\newblock Dust and reflection removal from videos captured in moving car.
\newblock In {\em 2016 IEEE 13th International Conference on e-Business Engineering (ICEBE)}, pages 182--187. IEEE, 2016.

\bibitem{jeon2024spectral}
Yujin Jeon, Eunsue Choi, Youngchan Kim, Yunseong Moon, Khalid Omer, Felix Heide, and Seung-Hwan Baek.
\newblock Spectral and polarization vision: Spectro-polarimetric real-world dataset.
\newblock In {\em Proceedings of the IEEE/CVF Conference on Computer Vision and Pattern Recognition}, pages 22098--22108, 2024.

\bibitem{johnson2016perceptual}
Justin Johnson, Alexandre Alahi, and Li Fei-Fei.
\newblock Perceptual losses for real-time style transfer and super-resolution.
\newblock In {\em European conference on computer vision}, pages 694--711. Springer, 2016.

\bibitem{kim2020single}
Soomin Kim, Yuchi Huo, and Sung-Eui Yoon.
\newblock Single image reflection removal with physically-based training images.
\newblock In {\em Proceedings of the IEEE/CVF Conference on Computer Vision and Pattern Recognition}, pages 5164--5173, 2020.

\bibitem{kong2013physically}
Naejin Kong, Yu-Wing Tai, and Joseph~S Shin.
\newblock A physically-based approach to reflection separation: from physical modeling to constrained optimization.
\newblock {\em IEEE transactions on pattern analysis and machine intelligence}, 36(2):209--221, 2013.

\bibitem{lei2021robust}
Chenyang Lei and Qifeng Chen.
\newblock Robust reflection removal with reflection-free flash-only cues.
\newblock In {\em Proceedings of the IEEE/CVF Conference on Computer Vision and Pattern Recognition}, pages 14811--14820, 2021.

\bibitem{lei2020polarized}
Chenyang Lei, Xuhua Huang, Mengdi Zhang, Qiong Yan, Wenxiu Sun, and Qifeng Chen.
\newblock Polarized reflection removal with perfect alignment in the wild.
\newblock In {\em Proceedings of the IEEE/CVF conference on computer vision and pattern recognition}, pages 1750--1758, 2020.

\bibitem{lei2023robust}
Chenyang Lei, Xudong Jiang, and Qifeng Chen.
\newblock Robust reflection removal with flash-only cues in the wild.
\newblock {\em IEEE Transactions on Pattern Analysis and Machine Intelligence}, 2023.

\bibitem{levin2007user}
Anat Levin and Yair Weiss.
\newblock User assisted separation of reflections from a single image using a sparsity prior.
\newblock {\em IEEE Transactions on Pattern Analysis and Machine Intelligence}, 29(9):1647--1654, 2007.

\bibitem{li2020single}
Chao Li, Yixiao Yang, Kun He, Stephen Lin, and John~E Hopcroft.
\newblock Single image reflection removal through cascaded refinement.
\newblock In {\em Proceedings of the IEEE/CVF Conference on Computer Vision and Pattern Recognition}, pages 3565--3574, 2020.

\bibitem{li2020improved}
Tingtian Li, Yuk-Hee Chan, and Daniel~PK Lun.
\newblock Improved multiple-image-based reflection removal algorithm using deep neural networks.
\newblock {\em IEEE Transactions on Image Processing}, 30:68--79, 2020.

\bibitem{li2023polarimetric}
Xiaobo Li, Lei Yan, Pengfei Qi, Liping Zhang, Fran{\c{c}}ois Goudail, Tiegen Liu, Jingsheng Zhai, and Haofeng Hu.
\newblock Polarimetric imaging via deep learning: A review.
\newblock {\em Remote Sensing}, 15(6):1540, 2023.

\bibitem{li2014single}
Yu Li and Michael~S Brown.
\newblock Single image layer separation using relative smoothness.
\newblock In {\em Proceedings of the IEEE conference on computer vision and pattern recognition}, pages 2752--2759, 2014.

\bibitem{lyu2019reflection}
Youwei Lyu, Zhaopeng Cui, Si Li, Marc Pollefeys, and Boxin Shi.
\newblock Reflection separation using a pair of unpolarized and polarized images.
\newblock {\em Advances in neural information processing systems}, 32, 2019.

\bibitem{nandoriya2017video}
Ajay Nandoriya, Mohamed Elgharib, Changil Kim, Mohamed Hefeeda, and Wojciech Matusik.
\newblock Video reflection removal through spatio-temporal optimization.
\newblock In {\em Proceedings of the IEEE International Conference on Computer Vision}, pages 2411--2419, 2017.

\bibitem{nayar1997separation}
Shree~K Nayar, Xi-Sheng Fang, and Terrance Boult.
\newblock Separation of reflection components using color and polarization.
\newblock {\em International Journal of Computer Vision}, 21(3):163--186, 1997.

\bibitem{niklaus2021learned}
Simon Niklaus, Xuaner~Cecilia Zhang, Jonathan~T Barron, Neal Wadhwa, Rahul Garg, Feng Liu, and Tianfan Xue.
\newblock Learned dual-view reflection removal.
\newblock In {\em Proceedings of the IEEE/CVF Winter Conference on Applications of Computer Vision}, pages 3713--3722, 2021.

\bibitem{schechner1999polarization}
Yoav~Y Schechner, Joseph Shamir, and Nahum Kiryati.
\newblock Polarization-based decorrelation of transparent layers: The inclination angle of an invisible surface.
\newblock In {\em Proceedings of the seventh IEEE international conference on computer vision}, volume~2, pages 814--819. IEEE, 1999.

\bibitem{schechner2000polarization}
Yoav~Y Schechner, Joseph Shamir, and Nahum Kiryati.
\newblock Polarization and statistical analysis of scenes containing a semireflector.
\newblock {\em JOSA A}, 17(2):276--284, 2000.

\bibitem{shih2015reflection}
YiChang Shih, Dilip Krishnan, Fredo Durand, and William~T Freeman.
\newblock Reflection removal using ghosting cues.
\newblock In {\em Proceedings of the IEEE conference on computer vision and pattern recognition}, pages 3193--3201, 2015.

\bibitem{sony_polarsens}
{Sony Semiconductor Solutions}.
\newblock Polarsens: Polarization sensing technology, 2024.
\newblock Accessed: 2024-11-15.

\bibitem{strong2003edge}
David Strong and Tony Chan.
\newblock Edge-preserving and scale-dependent properties of total variation regularization.
\newblock {\em Inverse problems}, 19(6):S165, 2003.

\bibitem{tan2005separating}
Robby~T Tan and Katsushi Ikeuchi.
\newblock Separating reflection components of textured surfaces using a single image.
\newblock {\em IEEE transactions on pattern analysis and machine intelligence}, 27(2):178--193, 2005.

\bibitem{wan2018crrn}
Renjie Wan, Boxin Shi, Ling-Yu Duan, Ah-Hwee Tan, and Alex~C Kot.
\newblock Crrn: Multi-scale guided concurrent reflection removal network.
\newblock In {\em Proceedings of the IEEE Conference on Computer Vision and Pattern Recognition}, pages 4777--4785, 2018.

\bibitem{wan2019corrn}
Renjie Wan, Boxin Shi, Haoliang Li, Ling-Yu Duan, Ah-Hwee Tan, and Alex~C Kot.
\newblock Corrn: Cooperative reflection removal network.
\newblock {\em IEEE transactions on pattern analysis and machine intelligence}, 42(12):2969--2982, 2019.

\bibitem{wang2023personalized}
Mengyi Wang, Xinxin Zhang, Yongshun Gong, and Yilong Yin.
\newblock Personalized single image reflection removal network through adaptive cascade refinement.
\newblock In {\em Proceedings of the 31st ACM International Conference on Multimedia}, pages 8204--8213, 2023.

\bibitem{wei2019single}
Kaixuan Wei, Jiaolong Yang, Ying Fu, David Wipf, and Hua Huang.
\newblock Single image reflection removal exploiting misaligned training data and network enhancements.
\newblock In {\em Proceedings of the IEEE/CVF Conference on Computer Vision and Pattern Recognition}, pages 8178--8187, 2019.

\bibitem{wieschollek2018separating}
Patrick Wieschollek, Orazio Gallo, Jinwei Gu, and Jan Kautz.
\newblock Separating reflection and transmission images in the wild.
\newblock In {\em Proceedings of the European Conference on Computer Vision (ECCV)}, pages 89--104, 2018.

\bibitem{wu2023q}
Haoning Wu, Zicheng Zhang, Weixia Zhang, Chaofeng Chen, Liang Liao, Chunyi Li, Yixuan Gao, Annan Wang, Erli Zhang, Wenxiu Sun, et~al.
\newblock Q-align: Teaching lmms for visual scoring via discrete text-defined levels.
\newblock {\em arXiv preprint arXiv:2312.17090}, 2023.

\bibitem{xue2015computational}
Tianfan Xue, Michael Rubinstein, Ce Liu, and William~T Freeman.
\newblock A computational approach for obstruction-free photography.
\newblock {\em ACM Transactions on Graphics (TOG)}, 34(4):1--11, 2015.

\bibitem{yao2024neural}
Mingde Yao, Ruikang Xu, Yuanshen Guan, Jie Huang, and Zhiwei Xiong.
\newblock Neural degradation representation learning for all-in-one image restoration.
\newblock {\em IEEE Transactions on Image Processing}, 2024.

\bibitem{yun2018reflection}
Jae-Seong Yun and Jae-Young Sim.
\newblock Reflection removal for large-scale 3d point clouds.
\newblock In {\em Proceedings of the IEEE Conference on Computer Vision and Pattern Recognition}, pages 4597--4605, 2018.

\bibitem{zhang2018single}
Xuaner Zhang, Ren Ng, and Qifeng Chen.
\newblock Single image reflection separation with perceptual losses.
\newblock In {\em Proceedings of the IEEE conference on computer vision and pattern recognition}, pages 4786--4794, 2018.

\bibitem{zhong2024language}
Haofeng Zhong, Yuchen Hong, Shuchen Weng, Jinxiu Liang, and Boxin Shi.
\newblock Language-guided image reflection separation.
\newblock In {\em Proceedings of the IEEE/CVF Conference on Computer Vision and Pattern Recognition}, pages 24913--24922, 2024.

\bibitem{zhu2024revisiting}
Yurui Zhu, Xueyang Fu, Peng-Tao Jiang, Hao Zhang, Qibin Sun, Jinwei Chen, Zheng-Jun Zha, and Bo Li.
\newblock Revisiting single image reflection removal in the wild.
\newblock In {\em Proceedings of the IEEE/CVF Conference on Computer Vision and Pattern Recognition}, pages 25468--25478, 2024.

\end{thebibliography}
}

\clearpage
\end{document}